%% file: arxiv_paper.tex
\documentclass[10pt,twocolumn,letterpaper]{article}

\usepackage[pagenumbers]{cvpr} 

\usepackage{graphicx}
\usepackage{amsmath}
\usepackage{amssymb}
\usepackage{booktabs}
\usepackage{array}
\usepackage{siunitx}
\usepackage{times}
\usepackage{makecell}

\usepackage[pagebackref,breaklinks,colorlinks]{hyperref}

\usepackage[capitalize]{cleveref}
\crefname{section}{Sec.}{Secs.}
\Crefname{section}{Section}{Sections}
\Crefname{table}{Table}{Tables}
\crefname{table}{Tab.}{Tabs.}

\newcolumntype{C}[1]{>{\centering\let\newline\\\arraybackslash\hspace{0pt}}m{#1}}
\newcolumntype{L}[1]{>{\raggedright\let\newline\\\arraybackslash\hspace{0pt}}m{#1}}
\begin{document}

\title{Synchronized Audio-Visual Frames with Fractional Positional Encoding for Transformers in Video-to-Text Translation}

\author{Philipp Harzig \quad Moritz Einfalt \quad Rainer Lienhart\\
University of Augsburg, Augsburg\\
{\tt\small \{firstname.lastname\}@uni-a.de}
}
\maketitle

\begin{abstract}
   Video-to-Text (VTT) is the task of automatically generating descriptions for short audio-visual video clips, which can support visually impaired people to understand scenes of a YouTube video for instance.
   Transformer architectures have shown great performance in both machine translation and image captioning, lacking a straightforward and reproducible application for VTT.
   However, there is no comprehensive study on different strategies and advices for video description generation including exploiting the accompanying audio with fully self-attentive networks.
   Thus, we explore promising approaches from image captioning and video processing and apply them to VTT by developing a straightforward Transformer architecture. Additionally, we present a novel way of synchronizing audio and video features in Transformers which we call Fractional Positional Encoding (FPE).
   We run multiple experiments on the VATEX dataset to determine a configuration applicable to unseen datasets that helps describing short video clips in natural language and improved the CIDEr and BLEU-4 scores by 37.13 and 12.83 points compared to a vanilla Transformer network and achieve state-of-the art results on the MSR-VTT and MSVD datasets. Also, FPE helps increase the CIDEr score by a relative factor of \SI{8.6}{\percent}.
\end{abstract}

\section{Introduction}

Recurrent Neural Networks are a common architecture to model language generation tasks. Especially Long short-term memory (LSTM) Networks in combination with Deep Convolutional Neural Networks are used to generate descriptions of images \cite{vinyals2015show, karpathy2015deep, johnson2016densecap}. The architectures have matured over the years and introduced self-attention for LSTM Layers~\cite{xu2015show}. 
These methods have also become more and more popular for machine translation tasks, whose encoder-decoder architecture originally inspired the Show and Tell model of Vinyals~et~al.~\cite{vinyals2015show}. Recently, Vaswani~et~al.~\cite{vaswani2017attention} introduced a simple network architecture that is solely based on attention mechanisms and gets rid of convolutions altogether. Given the massive improvements in the task of sequence transduction and machine translation, it is natural to adapt this technique to Image Captioning~\cite{cornia2020meshed}. 

In this work, we focus on the Video-to-Text (VTT) task, which is actually quite similar to Image Captioning. We develop a model that is easy to implement and yet generates high-quality captions. We start with a Transformer modified to cope with video inputs as baseline and investigate several improvements by adopting various techniques from the domain of Image Captioning. 
We focus on promising extensions in order to develop a model which is easy to reproduce.
Ultimately, we present a way to easily align video and audio features independent of their respective sampling rates. We align the features by extending the Positional Encoding to support fractional positions. \\
Our contributions are as follows:
\begin{itemize}
    \item We develop a simple Transformer model for generating descriptions for short video clips. We reuse and adopt promising approaches from Image Captioning and human action classification for video clips that does not consist of an ensemble of multiple models.
    \item We present a combination of learning rate schedules that increases performance and shortens convergence time for VTT. 
    \item Finally, we introduce Fractional Positional Encoding (FPE), an extension to the traditional Positional Encoding, which allows to synchronize video and audio frames dependent on their respective sampling rate. By using FPE, we improve our CIDEr score by 37.13 points in comparison to the baseline. Furthermore, we achieve state-of-the-art scores on the MSVD and MSR-VTT datasets.
\end{itemize}

\section{Related Work}
Generating captions automatically from images is a task that has been widely studied.
Most image captioning models are inspired by the machine translation encoder-decoder architecture and come with a vision CNN encoder and a language generating Recurrent Neural Network (RNN)~\cite{vinyals2015show, karpathy2015deep, donahue2015long}. Shortly after these inital works on Image Captioning, visual attention mechanisms have shown to benefit image description generation~\cite{xu2015show, anderson2018bottom}.

Video-to-Text (VTT) is the natural continuation to Image Captioning. Instead of generating short descriptions for still images, VTT tries to infer descriptions from short video clips. Pan et al.~\cite{pan2016jointly} use an encoder that utilizes 3D and 2D CNN features while the decoder is LSTM based. Many other works~\cite{gao2017video, gan2017semantic, gan2017stylenet} make use of 2D and/or 3D features in the encoder and generate the descriptions with an LSTM decoder. Similar to Image Captioning, works in VTT have adopted traditional attention mechanisms~\cite{long2018video, wang2018reconstruction, wang2018m3, pei2019memory, liu2020sibnet, zhang2019object, chen2019motion, wang2019controllable, hou2019joint, zhang2020object} and use object-level features~\cite{zhang2019object, aafaq2019spatio, zhang2020object, guo2020normalized} in the encoder to improve the generation of descriptions.

One big leap for machine translation was the introduction of the Transformer architecture by Vaswani et al.~\cite{vaswani2017attention}. By replacing recurrence with self-attention modules, they better utilized long-term dependencies and improved the state-of-the-art at a fraction of the training cost. 
Similar to the recurrent machine translation models, the Transformer architecture was quickly adopted in the task of image captioning~\cite{li2019entangled, cornia2020meshed, he2020image, yu2019multimodal}.

As Transformers operate on sequences of features, it is easy to modify this architecture to describe short video clips. Various other video description datasets depicting everyday activities have been presented~\cite{xu2016msr, 2020trecvidawad, pan2020auto, chen2011collecting}. In this work, we mainly focus on the VATEX Captioning dataset~\cite{lin2020multi}, which has also been used in the Video-to-Text (VTT) task~\cite{zhu2019vatex, lin2020multi, singh2020nits, zhang2020object, zhang2021open, guo2020normalized}. Furthermore, we validate our models on the MSR-VTT~\cite{xu2016msr} and MSVD~\cite{chen2011collecting} datasets.

\section{Model}
\label{sec:model}
We utilize a slightly modified Transformer~\cite{vaswani2017attention} as our baseline model.
The Transformer architecture is built around the idea of transforming sequences from one domain to another, i.e., the original Transformer is a machine translation model that operates on sequences of tokens (words). However, we work on a different input domain (i.e., video clips) instead of sentences. Thus, we modified the encoder of the original Transformer architecture by altering its inputs. For the baseline architecture, we feed the encoder with embedded images $\mathbf{x}=(x_1,\dots,x_n)$ for every video frame instead of embedded tokens.
After embedding the image features, we add the positional encoding on top of these embeddings in order to maintain information about absolute and relative ordering of the sequence. As videos are sequences of frames, we can adopt the same Positional Encoding that Vaswani et al.~\cite{vaswani2017attention} utilize for sequences of tokens.
Our baseline model has $N=8$ encoder layers and outputs continuous representations $\mathbf{z}=(z_1,\dots,z_n)$ of dimension $d_\textrm{model}=512$. Our decoder also has $N=8$ layers and generates an output sequence $\mathbf{y}=(y_1,\dots,y_m)$. We use a learned word embedding to convert the input tokens to vectors of dimension $d_\textrm{model}$ and share the weight matrix with a learned linear projection layer to predict the probabilities of the next word~\cite{vaswani2017attention, press2017using}. Given the embedded tokens and $\mathbf{z}$, the decoder generates its output one word $y_t$ at a time. Similar to most encoder-decoder sequence models, the decoder uses the output of the previous step as input to the current step in an auto-regressive way when generating text. Thus, we simply optimize the cross-entropy loss for every target sentence $S$ with words $S_t$ during training
\begin{equation}
    L(S) = - \sum^{m}_{t=1}\log y_t [S_t].
\end{equation}
Still, our adapted Transformer architecture is a model designed with Natural Language transduction in mind. Therefore, in order to gradually improve the baseline model, we employ techniques and methods from the related task of Image Captioning and adapt the architecture to use video clips as inputs.
Additionally, we introduce the novel Fractional Positional Encoding that allows to synchronize audio-visual frames in a Transformer encoder.
In the following, we motivate the changes made to the baseline architecture and preprocessing in order to improve the quality of the generated sentences.

\subsection{Memory-Augmented Encoder}
The weights learned for the self-attention layers only depend on pairwise similarities between the projected inputs, i.e., in our case the self-attention in the encoder only models pairwise relationships between single frames. This property of the self-attention in Transformers leads to a limitation. That is, we cannot memorize knowledge about relationships between frames that later help to describe contents of unseen videos.
To mitigate this issue, we make use of the memory-augmented encoding~\cite{cornia2020meshed}, which encodes multi-level visual relationships with a-priori knowledge. In the original work, Cornia et al.~\cite{cornia2020meshed} use a persistent, learnable memory vector which is concatenated to the key and value of the self-attention blocks of the Transformer's encoder.
These memory vectors allow to encode persistent a-priori knowledge about relationships between image regions. In contrast to~\cite{cornia2020meshed}, we work with video sequences instead of still images with regions. Adapted to our architecture,  we can encode prior knowledge about relationships between frames for each training video, which later can be transferred to unseen video samples.
\begin{figure*}[h!]
	\centering
	\includegraphics[width=\textwidth]{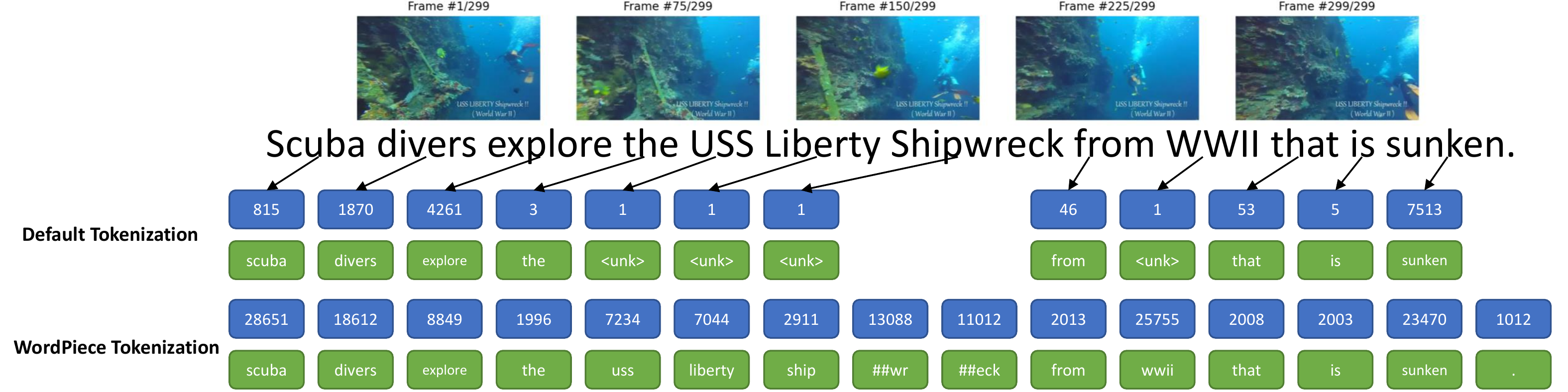}
	\caption{Example tokenization of a training sample from the VATEX dataset (Video ID: tXzq4vvGabk) with both default tokenization and WordPiece tokenization. WordPiece tokenization splits up rare words into subwords.}
	\label{fig:vocab_tokenization}
\end{figure*}
\subsection{Inflated 3D ConvNet}
A quick and naive way to implement a simple VTT model is to use frame-by-frame features extracted by a CNN designed.
However, some actions in a video clip can only be inferred by looking at temporal changes. A standard image-based CNN never gets to see what happens before or after a given frame and cannot understand these kinds of semantics, thus, is not able to reflect them in the image features. One way to extract features over multiple frames is using 3D convolutions that not only operate on the spatial dimensions but also convolve over the time dimension.\\
Therefore, we use the well-known Inflated 3D ConvNet (I3D)~\cite{carreira2017quo} architecture in order to provide the encoder with better input features. In particular, we extract features from the videos with the RGB-I3D model, which was pretrained on the Kinetics Human Action Video dataset~\cite{kay2017kinetics}.

\subsection{Subword and BERT Vocabulary}
For the baseline model, we implement an ordinary dictionary that takes the $n_{\textrm{Voc}}$ most frequent words into account, i.e., we order the words/tokens within the captions descending according to their frequency (e.g., \textit{a} and \textit{the} are the two most frequent words).
However, an ordinary dictionary is limited by its size, e.g., 12,000 words. Rare words that are important for understanding some sentences are completely left out and replaced by an \textit{\textless unk\textgreater} token. \\
A possible solution for this problem is the WordPiece~tokenizer~\cite{wu2016google}, which we employ in this work. The WordPiece tokenizer is a model optimized to maximize the language-model likelihood of the training data while minimizing the corpus size given a number of desired tokens. The goal is to represent rare words by splitting them up into word-pieces, which can later be recovered. For our vocabulary, we use the default BERT~\cite{devlin-etal-2019-bert} tokens. In Figure~\ref{fig:vocab_tokenization}, we depict a sample tokenization of one training sample. The default tokenization, which we limit to $n_{\textrm{Voc}}=12000$ words cannot tokenize rare words such as \textit{USS}, \textit{liberty}, \textit{shipwreck} or \textit{WWII}. When using the WordPiece tokenizer with the BERT dictionary, we see that all words can be represented with tokens. Especially, the rare word \textit{shipwreck} is split up into three subwords (\textit{ship}, \textit{\#\#wr}, \textit{\#\#eck}). The leading \textit{\#\#} indicate a split-up word and we can reconstruct the whole word \textit{shipwreck} from the three tokens.
\subsection{Learning-Rate Scheduling}
\label{sec:lr-sched}
Similar to \cite{vaswani2017attention}, we employ a learning rate schedule that linearly increases the learning rate for the first warmup\_steps training steps. After the warm-up phase, we decrease the learning rate proportionally to the inverse square root of the current step ($\textrm{it}$). We set warmup\_steps to $10,000$ in our use case. In the following, we call this \textit{schedule-default}:
\begin{equation}
    \eta = d_\textrm{model}^{-0.5}\cdot \min(\textrm{it}^{-0.5}, \textrm{it}\cdot\textrm{warmup\_steps}^{-1.5}).
\end{equation}
For our models, it stands out that the validation score increases during the warm-up phase and continues to increase for two to three epochs during the slow decay of \textit{schedule-default}. However, after that, the validation scores decrease continuously, i.e., our model starts to overfit after a short while. Since the validation scores are strongly dependent on the learning rate according to our observations, we want to prevent early overfitting by using a different learning rate schedule.

The SGDR (Stochastic Gradient Descent with Warm Restarts)~\cite{loshchilov2016sgdr} learning rate schedule is a promising approach, as it is applied successfully in other related works~\cite{liu2018show, ding2021cogview, gu2017empirical} and helps to improve scores while speeding up convergence.
Initially, we found this technique to harm our final scores, i.e., the Transformer network did not seem to initialize correctly. However, when combining this approach with a warm-up phase, we did notice some improvements over \textit{schedule-default}. Particularly, we find that the learning-rate restarts harm the performance, but the fast cosine decay helps our model to converge faster and with better scores (see Section~\ref{sec:exp-transformers-for-vtt}). We depict \textit{schedule-sgdr} alongside \textit{schedule-default} in Figure~\ref{fig:sgdr-vs-score}. After the warm-up phase, we decay the learning rate for $T_0=5$ epochs. Other parameters according to \cite{loshchilov2016sgdr} are $T_{mult}=1.0$
\subsection{Naïve fusion of audio and video features}
\label{sec:model-audio}
When generating descriptions from visual data of video clips, we can inherently only describe what we ``see''.  However, some of the content reflected in the associated captions can only be derived when we also inject information about what we ``hear''. Since the videos of the VATEX dataset are videos from YouTube, we are able to extract raw audio from these video clips. 

To turn these raw audio streams into usable information, we extract audio features with the VGGish~\cite{hershey2017cnn} architecture that yields features of dimension $\mathbb{R}^{n_a\times128}$. We forward these features through a dense audio embedding layer to match $d_{\textrm{model}}=512$. \\
Note, that the number of audio frames $n_a$ does not match the number of image frames or I3D frames, respectively. Therefore, we cannot sum image and audio features and use these as new encoder input features. 
Because the sequence lengths of image features or I3D features are varying, we cannot simply concatenate vision and audio features as the added positional encoding may signal the encoder that it receives a vision feature as input when in reality it is an audio feature.

As we still want to allow vision features to attend to audio features and vice versa, we assume a fixed starting position for all audio features, which we set greater than the maximum number of vision input features $N_v$ within the dataset. More specifically, we concatenate $n_v$ vision and $n_a$ audio features along the time dimension while adding the position embeddings for indices $[0,\dots,n_v-1,N_v+0,\dots,N_v+n_a-1]$ onto them.

\subsection{Fractional Positional Encoding}

We present a novel way of aligning vision and audio features within a Transformer model. Our I3D frames and audio frames are not synchronized: for vision features we extract single frames without resampling the video and audio is resampled to \SI{16}{\kilo \hertz}. 
Thus, an I3D frame at a given position represents a different timestamp for videos with different framerates. If we resampled all videos to the same framerate, we would still have no way of synchronizing the vision frames with the audio frames, as those sampling rates differ. In other words, the audio frame at a given position would not match the timestamp of the I3D frame at the same position.

\begin{figure}[h]
	\centering
	\includegraphics[width=\columnwidth]{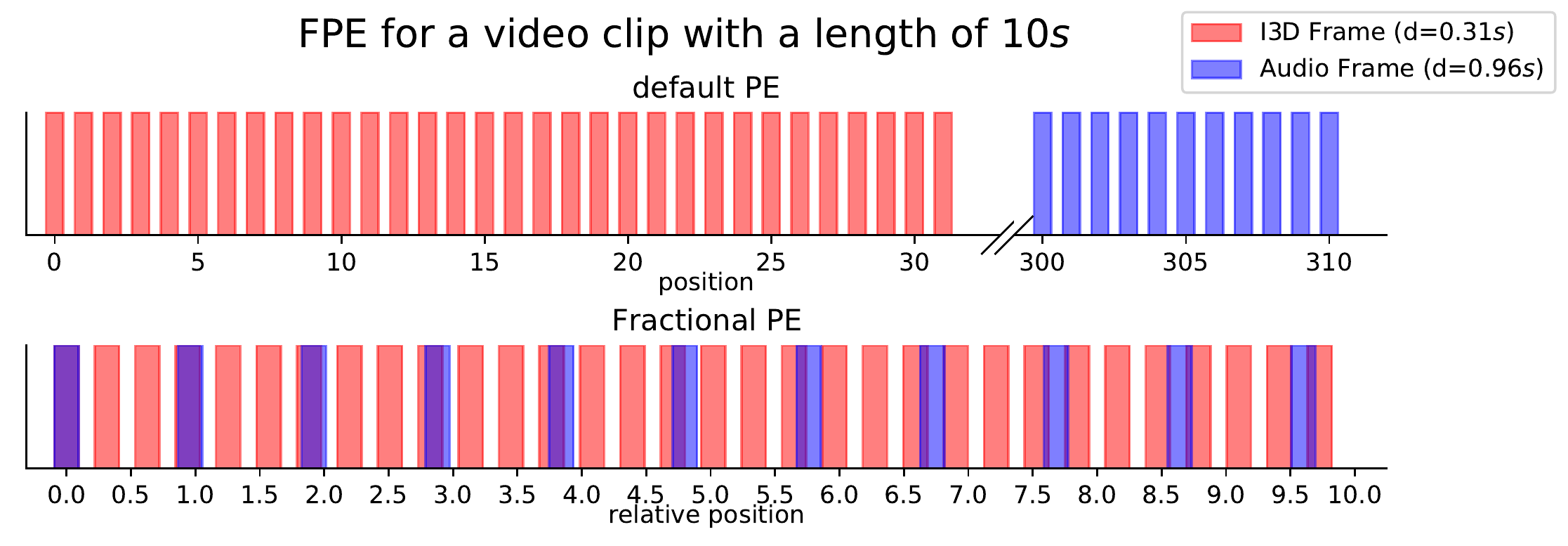}
	\caption{The default Positional Encoding for audio and video frames (on top) in comparison with the FPE (bottom) for an exemplary video. The video has $32$ I3D frames and $11$ audio frames. The lengths ($d$) of audio and video frames differ.}
	\label{fig:fpe}
\end{figure}

In the original work~\cite{vaswani2017attention}, the Positional Encoding has no inherent meaning other than to define the relative position of a word. For our input data however, vision and audio feature frames are aligned on the same time-axis and depend on their respective frame rate. Thus, we fix this problem by introducing the Fractional Positional Encoding (FPE) (see Figure~\ref{fig:fpe}). 
FPE is an extension to the traditional Positional Encoding that allows positional encoding on a fractional level. In order to fully utilize the audio features, the Transformer needs to know which audio frame corresponds to which vision frame. To do so, we calculate two timestamp factors for every video within the dataset, i.e., an audio and a vision timestamp factor. Both timestamp factors indicate the number of seconds each frame lasts. We then multiply the integer indices of each frame with the corresponding timestamp factor. Thus, we ensure that audio and video frames are properly aligned relative to their timestamp.

\subsection{Self-Critical Sequence Training}
\label{sec:scst}
In our baseline model, we optimize the objective of maximizing the likelihood of the next ground-truth word given previous ground-truth words and the encoder outputs. This approach is called ``Teacher-Forcing''~\cite{bengio2015scheduled} and has a serious drawback, i.e., the training phase is different from the inference phase (\textit{exposure bias}~\cite{ranzato2015sequence}). 
That is, during inference, we can only greedily sample the next word given previously sampled words. On ther hand, we maximize the probability of the next word given ground-truth words during training. 
In addition, our models are trained with cross-entropy loss and evaluated with non-differentiable metrics (e.g., CIDEr~\cite{vedantam2015cider} and BLEU~\cite{papineni2002bleu}).
In the work Self-critical Sequence Training (SCST) for Image Captioning, Rennie et al.~\cite{rennie2017self} present a sequence model that is trained to mitigate these problems.

SCST is a variation of the popular REINFORCE algorithm that utilizes the outputs of the model's test-time inference algorithm instead of estimating a baseline to directly optimize for the metric. 
The main idea of SCST is to baseline the reward with the reward of the current model in the test-time inference mode. Thus, we first greedily sample a caption for each video clip with our model in inference mode. Second, we sample $n_{\textrm{s}}=5$ sentences $w^s$ for the corresponding video clip in training mode using categorical sampling. 
Then, we calculate the CIDEr scores for the baseline caption $\hat{w}$ and the sampled captions $w^s$, respectively. Subsequently, we can baseline the reward of the sampled captions by subtracting the CIDEr score for the baseline caption.
Thus, sampled captions with a higher CIDEr score than the baseline caption get a positive reward and vice versa.
By optimizing for this objective, sampled captions with a higher CIDEr score will be increased in probability, while we try to make bad captions less likely. Note, that we assign the same reward to every word of each sampled caption. The gradient of the loss function can be approximated as follows:
\begin{equation}
    \nabla_{\theta}L(\theta) \approx -(r(w^s)-r(\hat{w})) \nabla_{\theta} \log_{p_{\theta}}(w^s).
\end{equation}
Each word will be weighted according to its $\log$ probability and $r(\cdot)$ is the reward function. $\theta$ are the parameters of the network and define a policy $p_{\theta}$.
For our final models, we additionally optimize the BLEU-4 metric. Therefore, our reward function becomes $r(\cdot) = \lambda_{\textrm{CIDEr}} \cdot r_{\textrm{CIDEr}}(\cdot) + \lambda_{\textrm{BLEU-4}}$
where $\lambda_{\cdot}$ is a weight for the corresponding metric.

\section{Datasets}
\label{sec:datasets}

We mainly use the VATEX Dataset~\cite{wang2019vatex} for our experiments.
The VATEX dataset is split into 4 sets, i.e., the training set, the validation set, the public test set and the private test set. The VATEX dataset comes with 10 English and 10 Chinese captions per video clip. Most video clips have a length of \SI{10}{\second}.
Additionally, we train our final models on the MSR-VTT~\cite{xu2016msr} and MSVD~\cite{chen2011collecting} datasets. For MSVD, we follow the common practice and split the \num{1970} available video clips into three partitions of \num{1200}, \num{100} and \num{670} for training, validation and test, respectively. For MSR-VTT, we use splits containing \SI{90}{\percent}, \SI{5}{\percent} and \SI{5}{\percent}.
We depict details about the datasets in Table~\ref{tab:datasets}.

\begin{table}[]
	\resizebox{\columnwidth}{!}{
	\begin{tabular}{@{}lllll@{}}
		\toprule
		Dataset & \# Videos (clips) & \# Sentences & \# Videos avail. & \# Sentences usable \\ \midrule
		VATEX~\cite{wang2019vatex}          & 41,269          & 349,910     & 38,109       & 323,950          \\
		MSR-VTT~\cite{xu2016msr}           & 10,000           & 200,000     & 7,773         & 155,460          \\
		MSVD~\cite{chen2011collecting}           & 2,089           & 85,550     & 1,970         & 80,838          \\
		\bottomrule
	\end{tabular}
	}
	\caption{Different datasets and their respective number of video clips and number of available videos. Sentences are available for every video, however, not every video was available to be downloaded from YouTube.}
	\label{tab:datasets}
\end{table}

\subsection{Preprocessing of Videos}

\textbf{Single Images.}
In order to process the videos in our model, we extract every frame of each video. We do not resample the videos to a fixed frame rate. We use ResNet-101~V2~\cite{he2016identity} to compute features for the extracted frames by resizing the input images to $224\times 224$ and using the average pooled features with dimension $\mathbb{R}^{n_v\times2048}$, where $n_v$ is the number of frames of the corresponding video clip.  \\ 
\textbf{I3D Features.} We extract I3D features similar to frame-level features. Instead of forwarding frame images through the ResNet-101~V2 network, we extract video clip features with the RGB-I3D pretrained on the Kinetics Human Action Video dataset~\cite{kay2017kinetics}. The I3D yields features of dimension $\mathbb{R}^{n_v\times 1024}$, whereas in this case $n_v$ is the number of I3D frames, which is less than the original number of frames in the video due to the 3D-convolutions. Furthermore, we average-pool over the spatial dimensions and repurpose the same image embedding layer to embed the I3D features into the model dimension ($\mathbb{R}^{n_v\times 512}$).  \\ 
\textbf{Audio features.} We take the audio of the video, resample it to \SI{16}{\kilo\hertz} and extract features with the VGGish~\cite{hershey2017cnn} network. This network yields features of dimension $\mathbb{R}^{n_a\times 128}$. Here, $n_a$ is the number of audio features, which is different from $n_v$. If no audio stream for a video is existent, we create a dummy feature vector with all zeros and dimension $\mathbb{R}^{1\times 128}$.

\subsection{Preprocessing of Tokens}
For our models with the default vocabulary, we employ a simple text tokenizer that filters out special characters\footnote{!"\#\textdollar{}\%\&()*+.,-/:;=?@[\textbackslash{}]\^{}\_`\{$\vert$\}\textasciitilde{}}. We limit the vocabulary to $12000$ tokens and replace less occurring words with the \textit{\textless unk\textgreater} token. \\
We tokenize ground-truth words with the WordPiece tokenizer with the English BERT vocabulary for our other models. Additionally, for a subset of our models, we load pretrained embedding weights\footnote{\url{https://tfhub.dev/google/small_bert/bert_uncased_L-8_H-512_A-8/1}} from the BERT\textsubscript{SMALL} model, which has $d_{\textrm{model}}=512$ to match our architecture.

\section{Experiments}
In our experimental part of this work, we evaluate the suggested extensions. We also present an extensive ablative study of our simple extensions from Section~\ref{sec:model} and show their effectiveness. At the end of the Section, we show that FPE improves scores substantially.
\begin{table*}[]
\resizebox{\textwidth}{!}{
\begin{tabular}{@{}ccccccc|ccccccc@{}}
\toprule
\textbf{Model} & \textbf{Features} & \textbf{$|\textrm{mv}|$} & \textbf{Vocabulary} & \textbf{ft We} & \textbf{FPE} & \textbf{lr Schedule} & \textbf{B@1} & \textbf{B@2} & \textbf{B@3} & \textbf{B@4} & \textbf{M} & \textbf{R} & \textbf{C} \\ \midrule
\textbf{baseline} & R101 & 0 & Default & \checkmark & --- & Default & 65.49 & 47.42 & 34.06 & 23.47 & 18.72 & 43.60 & 33.72 \\
\textbf{memvec} & R101 & 64 & Default & \checkmark & --- & Default & 66.56 & 48.40 & 35.08 & 24.57 & 18.83 & 43.92 & 35.23 \\
\textbf{i3d-baseline} & I3D & 0 & Default & \checkmark & --- & Default & 69.64 & 52.49 & 38.94 & 27.92 & 20.85 & 46.33 & 49.13 \\
\textbf{i3d-memvec} & I3D & 64 & Default & \checkmark & --- & Default & 71.82 & 54.51 & 40.63 & 29.21 & 21.77 & 47.33 & 53.01 \\
\textbf{i3d-wp} & I3D & 64 & WP & \checkmark & --- & Default & 71.36 & 54.24 & 40.50 & 29.04 & 21.49 & 47.08 & 50.19 \\
\textbf{i3d-wp-audio} & I3D+VGGish & 64 & WP & \checkmark & --- & Default & 72.20 & 55.64 & 41.68 & 29.99 & 22.11 & 48.10 & 52.64 \\
\textbf{i3d-audio} & I3D+VGGish & 64 & Default & \checkmark & --- & Default & 73.09 & 56.26 & 42.08 & 30.40 & 22.04 & 48.08 & 51.72 \\
\textbf{i3d-bert} & I3D & 64 & WP-BERT & --- & --- & Default & 71.12 & 53.83 & 39.79 & 28.26 & 21.76 & 47.14 & 51.38 \\
\textbf{i3d-bert-audio} & I3D+VGGish & 64 & WP-BERT & --- & --- & Default & 71.39 & 55.03 & 41.56 & 30.35 & 22.06 & 47.90 & 53.16 \\
\textbf{i3d-bert-ft-audio} & I3D+VGGish & 64 & WP-BERT & \checkmark & --- & Default & 72.64 & 55.97 & 42.04 & 30.32 & 21.95 & 48.09 & 52.09 \\
\textbf{i3d-bert-audio-sgdr} & I3D+VGGish & 64 & WP-BERT & --- & --- & sgdr & 73.53 & 57.55 & 43.81 & 32.16 & 22.70 & 49.07 & 56.92 \\
\textbf{i3d-bert-audio-sgdr-FPE} & I3D+VGGish & 64 & WP-BERT & --- & \checkmark & sgdr & 75.35 & 58.58 & 44.30 & 32.43 & 23.81 & 49.60 & 61.80 \\ \midrule
\textbf{SCST-Cider} & I3D+VGGish & 64 & WP-BERT & --- & --- & $5\cdot10^{-6}$ & 73.52 & 59.20 & 41.11 & 28.73 & 23.08 & 48.20 & 68.87 \\
\textbf{SCST-Cider-B4} & I3D+VGGish & 64 & WP-BERT & --- & --- & $5\cdot10^{-6}$ & 78.21 & 61.03 & 46.26 & 33.92 & 23.65 & 49.91 & 68.62 \\
\textbf{SCST-Cider-B4-FPE} & I3D+VGGish & 64 & WP-BERT & --- & \checkmark & $5\cdot10^{-6}$ & 78.74 & 62.82 & 48.64 & 36.30 & 24.52 & 51.91 & 70.85 \\ \bottomrule
\end{tabular}
}
\caption{Ablation study for our VTT Transformer models on the VATEX validation set. On the left, we list the model names with their respective configurations (ft We=fine-tune word embedding). On the right we list the validation scores (B@x=BLEU-x, M = METEOR, R = Rouge-L, C = CIDEr).}
\label{tab:val-results}
\end{table*}
\subsection{Implementation details}
Our model is implmented with TensorFlow 2 and we publish our code on GitHub\footnote{\url{https://github.com/fpe-vtt/ftt-vpe}}.\\
As a baseline model, we implement a vanilla Transformer~\cite{vaswani2017attention} model with $d_{\textrm{model}}=512,d_{ff}=2048$. Our encoder and decoder each have $N=8$ layers with $h=8$ parallel attention heads. We also adopt the same learning rate schedule from \cite{vaswani2017attention}, however, we change the number of warm-up steps to 10,000. 
As optimizer, we use Adam~\cite{kingma2014adam} with the learning rate schedules from Section~\ref{sec:lr-sched} and $\beta_1=0.9, \beta_2=0.999$ and $\epsilon=1\cdot 10^{-8}$. We train for a maximum number of 50 epochs with a batch size of 128 and employ early stopping based on the validation CIDEr score. 
Because of the huge memory demand of the SCST training, we lower the effective batch size to 16 (i.e. 4 GPUs with batch size 4) during fine-tuning stage and use a constant learning rate of $\eta=5\cdot10^{-6}$. 
\begin{figure}[h!]
	\centering
	\includegraphics[width=\columnwidth]{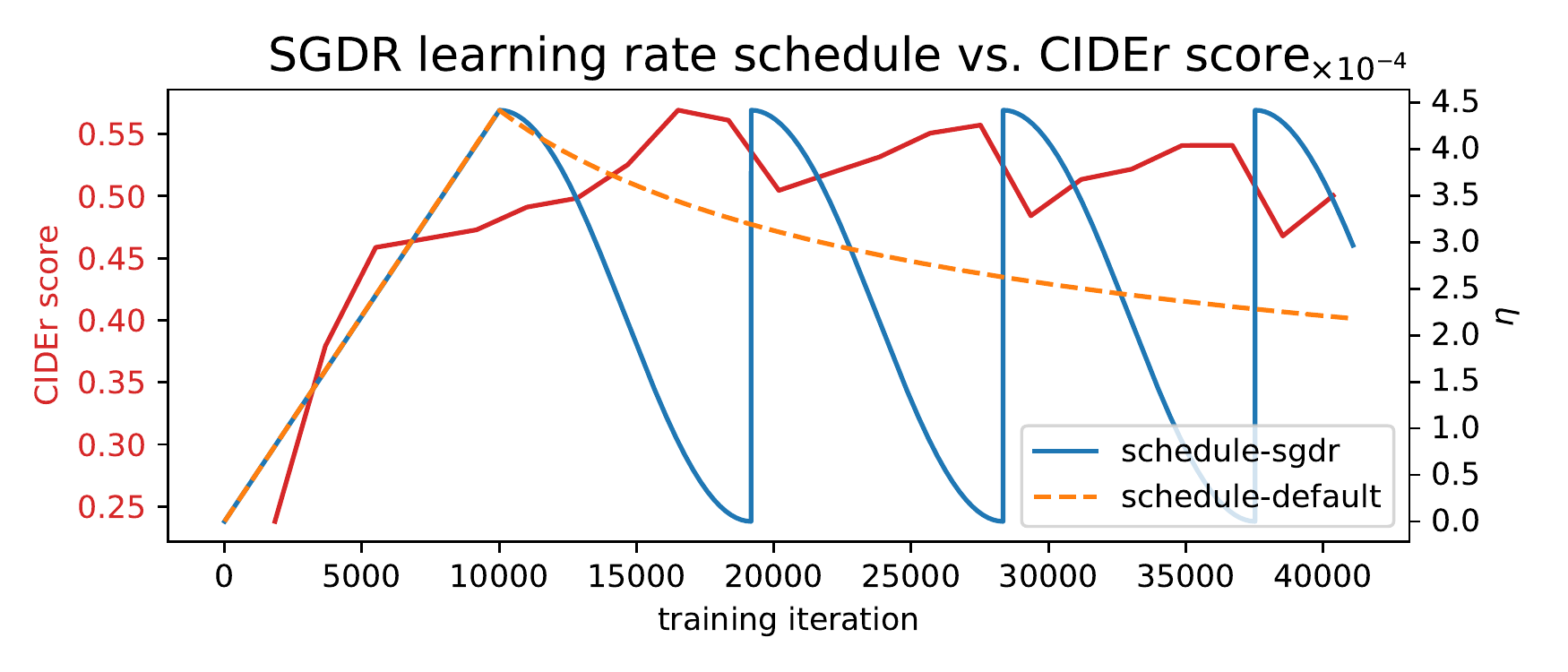}
	\caption{Course of learning rate plotted against the CIDEr validation score of model \textit{i3d-bert-audio-sgdr}. We plotted \textit{schedule-default} against \textit{schedule-sgdr} for comparison.}
	\label{fig:sgdr-vs-score}
\end{figure}

\begin{figure*}

\begin{tabular}{C{.64\linewidth} L{.36\linewidth}}
        \includegraphics[width=\linewidth]{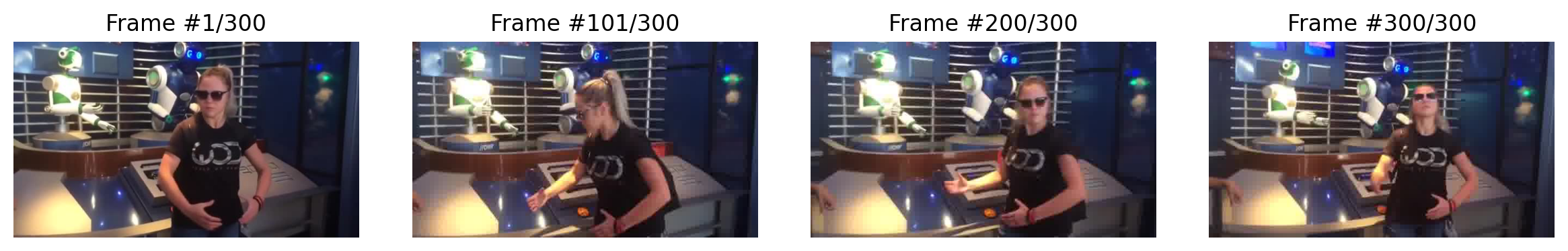} & {\tiny \input{generated_caps/wacv_sample_video_VATEX_val_Cx6XQIwOqyM_000002_000012_caps.txt}} \\ \midrule
        \includegraphics[width=\linewidth]{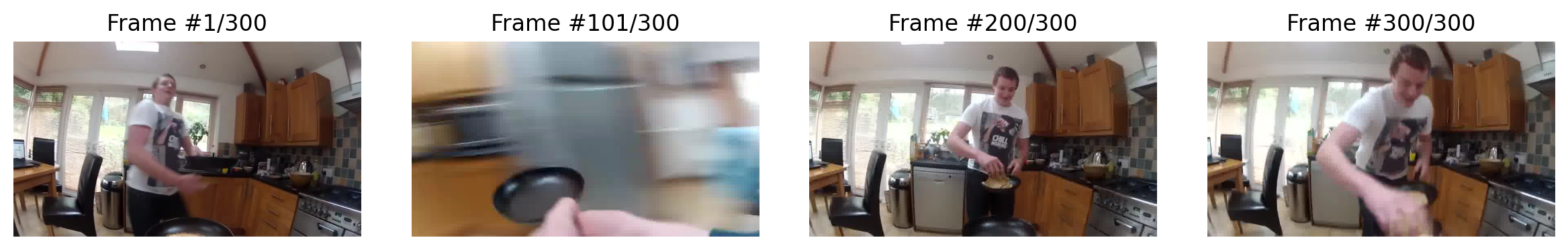} & {\tiny \input{generated_caps/wacv_sample_video_VATEX_val_9XGSi2nIY9E_000040_000050_caps.txt}} \\ \midrule
        \includegraphics[width=\linewidth]{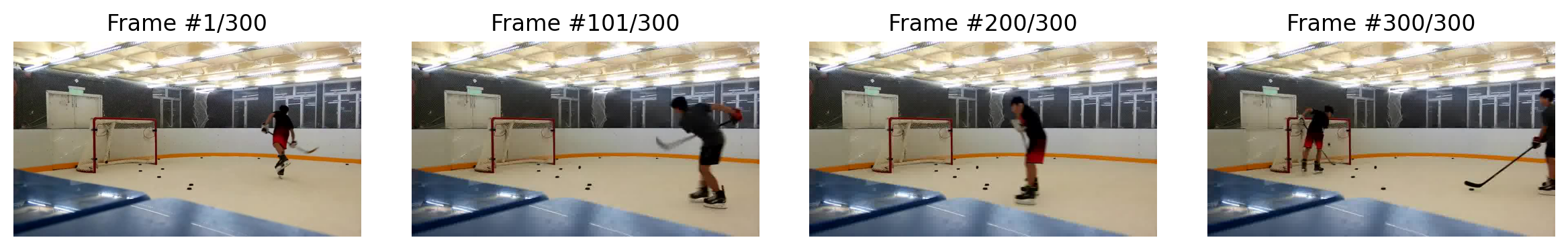} & {\tiny \input{generated_caps/wacv_sample_video_VATEX_val_DH_uqMhdyEk_000117_000127_caps.txt}}
\end{tabular}
\caption{Examples of generated descriptions for three example videos from the validation split. We see four frames from each video together with the frame number on the left and the generated caption for each model on the right.}
\label{fig:generated_example}
\end{figure*}
\subsection{Discussion of Results}
\label{sec:exp-transformers-for-vtt}
In the following, we discuss the results of the extensions presented in Section~\ref{sec:model}.
In Table~\ref{tab:val-results}, we depict results on the validation set of the VATEX dataset. In Figure~\ref{fig:generated_example}, we show generated captions for three example videos and every model from the VATEX validation set.
Both when looking at the scores and the generated descriptions, we see that our \textit{baseline} model scores worst across all metrics. 
The baseline model uses frame-level ResNet-101 features with an image embedding for the encoder.  \\ 
\textbf{Memory-Augmented Encoder.} Adding a memory vector to the key and value of the multi-head self-attention allows the encoder network to learn a-priori knowledge about relationships on an intra frame level. For example, when we look at sentences generated for the bottom video clip in Figure~\ref{fig:generated_example}, we see an ice hockey player doing some shots on a goal. Comparing the captions generated by the \textit{*memvec} models to the captions of the baseline models, we see the models have memorized that ice hockey often is played within an \textit{ice rink}.
In addition, for the model \textit{memvec}, we see a slight boost in all scores except Rouge-L, e.g., the CIDEr score is improved by about 1.5 points.  \\ 
\textbf{Image Features and I3D Features.} One of the two extensions gaining the most in terms of CIDEr score is replacing frame-level image features with features from the RGB-I3D network. Looking at the model \textit{i3d-baseline} that uses no memory-augmented encoder, we see an increase of 15.41 points and 5.39 points in CIDEr and BLEU-4, respectively. The memory-augmented encoder benefits from the I3D features in the same way, i.e., \textit{i3d-memvec} gains 16.63 points and 5.12 points in CIDEr and BLEU-4, respectively.  \\ 
\textbf{Naïve fusion of audio and video features.} Most videos not only contain visual data but also audio data. Thus, it is obvious that some contents of a textual description of said video clip can only be described accurately when also using the audio data. 
As already described in Section~\ref{sec:model-audio}, we first concatenate vision and audio features and naively add traditional positional encodings onto them.
When comparing the model \textit{i3d-audio} with \textit{i3d-memvec} we can observe no gains in performance, i.e., the CIDEr score is slightly worse with -0.14 points while BLEU-4 improves the score by 0.71, However, we will see in the next paragraph that combining audio features with the BERT dictionary will yield improvements.  \\ 
\textbf{Dictionaries and Tokenization.} We already stated that by using WordPiece tokenization, we are able to generate rare words that do not occur in a vocabulary with fixed size. However, this does not necessarily reflect on the scores as model \textit{i3d-wp} shows. In comparison to \textit{i3d-memvec} we loose 1.6 points and 0.66 points in CIDEr and BLEU-4, respectively. Similarly, when initializing the word embedding with BERT embeddings (\textit{i3d-bert}), we loose 0.48 points and 0.72 points for CIDEr and BLEU-4, respectively. However, in combination with the concatenated audio features WordPiece tokenization gives us better results. For WordPiece tokenization with no initialization of the word embedding, we gain 0.77 points (CIDEr) and 1.16 points (B-4). When initializing the word embeddings (\textit{i3d-bert-audio}) we get slightly higher scores. When comparing \textit{i3d-bert-audio} with \textit{i3d-bert}, we also see the benefit of audio features, which could not be seen beforehand. Note, that we keep the BERT embeddings frozen during training, because fine-tuning them hurts performance (see model \textit{i3d-bert-ft-audio}). \\ 
\textbf{Learning Rate Scheduling.} Replacing the default Transformer learning rate schedule with our modified version of SGDR \textit{i3d-bert-audio-sgdr} improves the performance by 3.76 points and 1.81 points in CIDEr and BLEU-4, respectively. As we have already discussed in Section~\ref{sec:lr-sched}, the fast decay of the SGDR schedule helps to boost our validation scores as we depict in Figure~\ref{fig:sgdr-vs-score}. After the warm-up phase of 10,000 steps, the validation accuracy makes another climb until it hits its maximum CIDEr score of 56.92 at the end of the first decay. However, we see that restarting the learning rate leads to a drop in performance. Our model can recover somewhat but never reaches the maximum score again and its performance declines slowly.  \\ 
\begin{table*}[]
\resizebox{\textwidth}{!}{
\begin{tabular}{@{}cc|cccc|cccc|cccc|cccc@{}}
\toprule
\textbf{} & \textbf{} & \multicolumn{4}{c}{\textbf{Features}} & \multicolumn{4}{c}{\textbf{MSVD}} & \multicolumn{4}{c}{\textbf{MSR--VTT}} & \multicolumn{4}{c}{\textbf{VATEX}} \\ 
\multicolumn{1}{c}{\textbf{Model}} & \multicolumn{1}{c}{\textbf{Year}} & \textbf{I} & \textbf{M} & \textbf{O} & \textbf{A} & \textbf{B@4} & \textbf{M} & \textbf{R} & \textbf{C} & \textbf{B@4} & \textbf{M} & \textbf{R} & \textbf{C} & \textbf{B@4} & \textbf{M} & \textbf{R} & \textbf{C} \\ \midrule
\textbf{M3} & CVPR 2018~\cite{wang2018m3} & \checkmark & \checkmark & --- & --- & 52.8 & 33.3 & --- & --- & 38.1 & 26.6 & --- & --- & --- & --- & --- & --- \\
\textbf{RecNet} & ICCV 2018~\cite{wang2018reconstruction} & \checkmark & --- & --- & --- & 52.3 & 34.1 & 69.8 & 80.3 & 39.1 & 26.6 & 59.3 & 42.7 & --- & --- & --- & --- \\
\textbf{PickNet} & ECCV 2018~\cite{chen2018less} & \checkmark & --- & --- & --- & 52.3 & 33.3 & 69.6 & 76.5 & 41.3 & 27.7 & 59.8 & 44.1 & --- & --- & --- & --- \\
\textbf{MARN} & CVPR 2019~\cite{pei2019memory} & \checkmark & \checkmark & --- & --- & 48.6 & 35.1 & 71.9 & 92.2 & 40.4 & 28.1 & 60.7 & 47.1 & --- & --- & --- & --- \\
\textbf{SibNet} & ACM'MM 2019~\cite{liu2020sibnet} & \checkmark & --- & --- & --- & 54.2 & 34.8 & 71.7 & 88.2 & 40.9 & 27.5 & 60.2 & 47.5 & --- & --- & --- & --- \\
\textbf{OA-BTG} & CVPR 2019~\cite{zhang2019object} & \checkmark & --- & \checkmark & --- & 56.9 & 36.2 & --- & 90.6 & 41.4 & 28.2 & --- & 46.9 & --- & --- & --- & --- \\
\textbf{GRU-EVE} & CVPR 2019~\cite{aafaq2019spatio} & \checkmark & \checkmark & \checkmark & --- & 47.9 & 35 & 71.5 & 78.1 & 38.3 & 28.4 & 60.7 & 48.1 & --- & --- & --- & --- \\
\textbf{MGSA} & AAAI 2019~\cite{chen2019motion} & \checkmark & \checkmark & --- & \checkmark & 53.4 & 35 & --- & 86.7 & 42.4 & 27.6 & --- & 47.5 & --- & --- & --- & --- \\
\textbf{POS+CG} & CVPR 2019~\cite{wang2019controllable} & \checkmark & \checkmark & --- & --- & 52.5 & 34.1 & 71.3 & 88.7 & 42 & 28.2 & 61.6 & 48.7 & --- & --- & --- & --- \\
\textbf{POS+VCT} & ICCV 2019~\cite{hou2019joint} & \checkmark & \checkmark & --- & --- & 52.8 & 36.1 & 71.8 & 87.8 & 42.3 & 29.7 & 62.8 & 49.1 & --- & --- & --- & --- \\
\textbf{ORG-TRL} & CVPR 2020~\cite{zhang2020object} & \checkmark & \checkmark & \checkmark & --- & 54.3 & 36.4 & 73.9 & 95.2 & 43.6 & 28.8 & 62.1 & 50.9 & 32.1 & 22.2 & 48.9 & 49.7 \\
\textbf{LSTM-TSA\textsubscript{IV}} & CVPR 2017~\cite{pan2017video} &  &  &  &  & 52.8 & 33.5 & --- & --- & --- & --- & --- & --- & --- & --- & --- & --- \\
\textbf{aLSTMs} & IEEE ToM 2017~\cite{gao2017video} & \checkmark & \checkmark & --- & --- & 50.8 & 33.3 & --- & --- & 38 & 26.1 &  & 43.2 & --- & --- & --- & --- \\
\textbf{RCG} & CVPR 2021~\cite{zhang2021open} & \checkmark & \checkmark & --- & --- & --- & --- & --- & --- & 42.8 & 29.3 & 61.7 & 52.9 & 33.9 & 23.7 & 50.2 & 57.5 \\
\textbf{NSA} & CVPR 2020~\cite{guo2020normalized} & --- & \checkmark & \checkmark & --- & --- & --- & --- & --- & --- & --- & --- & --- & 31.4 & 22.7 & 49 & 57.1 \\
\textbf{SemSynAN} & CVPR 2021~\cite{perez2021improving} & \checkmark & \checkmark & --- & --- & \textbf{64.4} & \textbf{41.9} & \textbf{79.5} & \textbf{111.5} & \textbf{46.4} & \textbf{30.4} & \textbf{64.7} & 51.9 & --- & --- & --- & --- \\
\textbf{VATEX} & CVPR 2019~\cite{wang2019vatex} & --- & \checkmark & --- & --- & --- & --- & --- & --- & --- & --- & --- & --- & 28.7 & 21.9 & 47.2 & 45.6 \\\midrule
\textbf{SCST-Cider-B4-FPE\footnotemark} & Ours & --- & \checkmark & --- & \checkmark & 51.22 & 34.73 & 72.69 & 103.2 & 45.91 & 30.25 & 64.12 & \textbf{62.11} & 33.28 & 22.74 & 49.56 & 54.63 \\ 
\midrule
\multicolumn{18}{l}{Non per-reviewed papers:}\\
\textbf{MV+HR} & arXiv 2019~\cite{zhu2019vatex} & \checkmark & \checkmark & \checkmark & ---  & --- & --- & --- & --- & --- & --- & --- & --- & \textbf{40.7} & \textbf{25.8} & \textbf{53.7} & \textbf{81.4} \\
\textbf{MM-Feat} & arXiv 2020~\cite{lin2020multi} & \checkmark & \checkmark & \checkmark & \checkmark & --- & --- & --- & --- & --- & --- & --- & --- & 39.2 & 26.5 & 52.7 & 76 \\
\textbf{NITS-VC} & arXiv 2020~\cite{singh2020nits} & --- & \checkmark & --- & --- & --- & --- & --- & --- & --- & --- & --- & --- & 22 & 18 & 43 & 27 \\ \bottomrule
\end{tabular}
}
\caption{Comparison on VATEX, MSVD and MSR-VTT datasets against state-of-the-art methods. For VATEX, we tested our model on the private test set with the evaluation server. For MSVD and MSR-VTT, we use the test-splits discussed in Section~\ref{sec:datasets}. I, M, O and A denote image, motion, object and audio features.}
\label{tab:test-results}
\end{table*}
\textbf{FPE.}
In contrast to the naïve fusion of audio and video features, FPE (\textit{i3d-bert-audio-sgdr-FPE}) boosts performance across all metrics significantly. Most notably, synchronizing audio and video features by their relative position has the hugest benefit on the CIDEr metric, where we gain 4.88 points. Even during self-critical fine tuning (see next paragraph), FPE (\textit{SCST-Cider-B4-FPE}) achieves improvements across all metrics. Thus, we conclude that FPE is an easy and effective way to synchronize audio and video features in Transformers. \\ 
\textbf{SCST.} We initialize the self-critical sequence training with the best models \textit{i3d-bert-audio-sgdr} and \textit{i3d-bert-audio-sgdr-FPE}. As reward function, we calculate the CIDEr score of the baseline caption and the sampled sentences. We see that directly optimizing the CIDEr metrics leads to big gains in the CIDEr metric, i.e., 68.87 points vs. 56.92 points. The difference of 11.95 points is the second biggest improvement besides replacing image features with I3D features. However, as we only optimize for the CIDEr metric in model \textit{SCST-Cider}, we loose 3.43 points on the BLEU-4 metric. We also loose some performance across all other metrics except Meteor. However, when directly optimizing for CIDEr and BLEU-4 (see model \textit{SCST-Cider-B4}; we set $\lambda_{\textrm{CIDEr}}=\lambda_{\textrm{BLEU-4}}=1.0$), we see that the CIDEr score is nearly identical while all BLEU-n scores get a significant boost. When combining SCST with FPE, our model produces the best results across all experiments and we improve by another 2.23 and 2.38 in CIDEr and BLEU-4, respectively.
\footnotetext[4]{We were only able to extract I3D and audio features for \num{5714}/\num{6278} video clips as the videos were no longer available on YouTube. We could use I3D features made available by the dataset's authors. These features, however, were different from our I3D features.}
\subsection{Comparison with State-of-the-Art}
We were not able to download all video files for the VATEX dataset from YouTube (see Table~\ref{tab:datasets}), thus we could not train, validate and test on the whole dataset. For the private test split of the VATEX dataset, we could download 5,714/6,278 videos, thus, missing features for 564 videos. However, the authors provide pre-extracted I3D features. After closer inspection, these features do not match our I3D features. Additionally, we do not have audio features for those missing videos. Submitting generated descriptions to the evaluation server requires descriptions for every single of the 6,278 videos, thus, we use the VATEX authors' I3D features with no audio features for submitting results. In Table~\ref{tab:test-results}, we depict results of our model \textit{SCST-Cider-B4-FPE} trained in the same manner on both train and validation splits. 
Our model scores not as well as the models from the VATEX video captioning challenge Zhu et al.~\cite{zhu2019vatex} and Lin et al.~\cite{lin2020multi}, who use ensembles of up to 32 models. However, across all published works on video captioning, we achieve similar performance on the reported metrics.
We also train our model on the MSVD and MSR-VTT datasets to prove the effectiveness of our method. On the MSVD dataset, our scores are below SemSynAN~\cite{perez2021improving} but otherwise better than all other methods listed in Table~\ref{tab:test-results}. For MSR-VTT, however, our final model outperforms SemSynAN by 10.21 points in CIDEr and performing similar to it for the other metrics.

\section{Future Work and Conclusion}
In our work, we presented a Transformer-based Video-to-Text architecture aimed to generate descriptions for short videos. Utilizing promising approaches from the related field of Image Captioning, we were able to gradually improve a vanilla Transformer designed for Machine Translation into a architecture that generates appropriate and matching captions for video clips. By combining motion features, audio features, a custom learning rate schedule and a pretrained vocabulary we establish a solid captioning model. Furthermore, we introduce the novel Fractional Positional Encoding to properly synchronize video and audio features with different sampling rates, which significantly improves results across all metrics. In combination with self-critical sequence training, we were able to considerably boost the performance of a baseline model by an absolute of 37.13 points or 210\% in the CIDEr metric. 

In the future, we want to expand our model with the X-Linear Attention block~\cite{pan2020x}, which shows huge potential in other works~\cite{zhu2019vatex}. We also want to improve our architecture further by employing unsupervised pretraining with the VideoBERT~\cite{sun2019videobert} model. 
Furthermore, we will extend the model by a multi-modal training objective that takes Chinese captions from the VATEX dataset into account in order to improve training feedback.

{\small
\bibliographystyle{ieee_fullname}
\bibliography{egbib}
}

\end{document}

%% file: generated_caps/wacv_sample_video_VATEX_val_Cx6XQIwOqyM_000002_000012_caps.txt
\makecell{\textbf{baseline:} a man is standing in a restaurant and talking about it \\
\textbf{memvec:} a man is sitting at a table playing a drum set \\
\textbf{i3d-baseline:} a woman is standing in front of a faucet and she is holding a bottle \\
\textbf{i3d-memvec:} a young boy is playing with a machine and a woman is talking \\
\textbf{i3d-wp:} a young man is standing at a ball game and talking to the camera . \\
\textbf{i3d-wp-audio:} a woman is standing in front of a store and she is talking to a man . \\
\textbf{i3d-audio:} a young boy is standing at a counter and he is standing in a chair \\
\textbf{i3d-bert:} a man is standing in front of a car and he is pumping gas in the air . \\
\textbf{i3d-bert-audio:} a young girl is at a bar and she is playing the game \\
\textbf{i3d-bert-audio-sgdr:} a young girl is trying to get her balance on the machine . \\
\textbf{i3d-bert-audio-sgdr-FPE:} a young girl is playing a toy game while a man watches her . \\
\textbf{SCST-Cider-B4:} a man and a young man is playing with a football player . \\
\textbf{SCST-Cider-B4-FPE:} a young girl is using a machine to peel a game of meat .}

%% file: generated_caps/wacv_sample_video_VATEX_val_9XGSi2nIY9E_000040_000050_caps.txt
\makecell{\textbf{baseline:} a man is flipping a pancake in a pan and catches it \\
\textbf{memvec:} a woman is sitting at a desk and talking about it \\
\textbf{i3d-baseline:} a man flips a pancake in the air and catches it \\
\textbf{i3d-memvec:} a woman is flipping a pancake in a pan and then she flips it \\
\textbf{i3d-wp:} a man is flipping a pancake in a frying pan . \\
\textbf{i3d-wp-audio:} a man is flipping a pancake in the air and catches it . \\
\textbf{i3d-audio:} a young girl is flipping a pancake and flipping it \\
\textbf{i3d-bert:} a man flips a pancake in a kitchen and flips it . \\
\textbf{i3d-bert-audio:} a man is flipping a pancake in a pan . \\
\textbf{i3d-bert-audio-sgdr:} a man is flipping a pancake in a pan . \\
\textbf{i3d-bert-audio-sgdr-FPE:} a man is flipping a pancake in a pan and then catches it . \\
\textbf{SCST-Cider-B4:} a man and a man flips a pancake and then in a frying pan . \\
\textbf{SCST-Cider-B4-FPE:} a young man is flipping a pancake in a frying pan .}

%% file: generated_caps/wacv_sample_video_VATEX_val_DH_uqMhdyEk_000117_000127_caps.txt
\makecell{\textbf{baseline:} a group of people are playing a game of curling \\
\textbf{memvec:} a man is playing a game of curling in a rink \\
\textbf{i3d-baseline:} a hockey player is skating backwards and then turns to a stop \\
\textbf{i3d-memvec:} a group of people are practicing skating on an ice rink \\
\textbf{i3d-wp:} a person is skating on an ice rink and practicing ice skating . \\
\textbf{i3d-wp-audio:} a group of people are playing hockey in an arena . \\
\textbf{i3d-audio:} a hockey game is being played on an ice rink \\
\textbf{i3d-bert:} a group of people are skating around in a hockey rink . \\
\textbf{i3d-bert-audio:} a group of people are playing hockey in a gym . \\
\textbf{i3d-bert-audio-sgdr:} a group of people are playing hockey in a rink . \\
\textbf{i3d-bert-audio-sgdr-FPE:} a group of people are playing a game of soccer in a indoor rink . \\
\textbf{SCST-Cider-B4:} a man and a person is playing a hockey goal on an ice rink . \\
\textbf{SCST-Cider-B4-FPE:} a group of people are playing a game of hockey on a rink .}

%% file: arxiv_paper.bbl
\begin{thebibliography}{10}\itemsep=-1pt

\bibitem{aafaq2019spatio}
Nayyer Aafaq, Naveed Akhtar, Wei Liu, Syed~Zulqarnain Gilani, and Ajmal Mian.
\newblock Spatio-temporal dynamics and semantic attribute enriched visual
  encoding for video captioning.
\newblock In {\em Proceedings of the IEEE/CVF Conference on Computer Vision and
  Pattern Recognition}, pages 12487--12496, 2019.

\bibitem{anderson2018bottom}
Peter Anderson, Xiaodong He, Chris Buehler, Damien Teney, Mark Johnson, Stephen
  Gould, and Lei Zhang.
\newblock Bottom-up and top-down attention for image captioning and visual
  question answering.
\newblock In {\em Proceedings of the IEEE conference on computer vision and
  pattern recognition}, pages 6077--6086, 2018.

\bibitem{2020trecvidawad}
George Awad, Asad~A. Butt, Keith Curtis, Yooyoung Lee, Jonathan Fiscus, Afzal
  Godil, Andrew Delgado, Jesse Zhang, Eliot Godard, Lukas Diduch, Jeffrey Liu,
  Alan~F. Smeaton, Yvette Graham, Gareth J.~F. Jones, Wessel Kraaij, and
  Georges Quénot.
\newblock Trecvid 2020: comprehensive campaign for evaluating video retrieval
  tasks across multiple application domains.
\newblock In {\em Proceedings of TRECVID 2020}. NIST, USA, 2020.

\bibitem{bengio2015scheduled}
Samy Bengio, Oriol Vinyals, Navdeep Jaitly, and Noam Shazeer.
\newblock Scheduled sampling for sequence prediction with recurrent neural
  networks.
\newblock {\em arXiv preprint arXiv:1506.03099}, 2015.

\bibitem{carreira2017quo}
Joao Carreira and Andrew Zisserman.
\newblock Quo vadis, action recognition? a new model and the kinetics dataset.
\newblock In {\em proceedings of the IEEE Conference on Computer Vision and
  Pattern Recognition}, pages 6299--6308, 2017.

\bibitem{chen2011collecting}
David Chen and William~B Dolan.
\newblock Collecting highly parallel data for paraphrase evaluation.
\newblock In {\em Proceedings of the 49th annual meeting of the association for
  computational linguistics: human language technologies}, pages 190--200,
  2011.

\bibitem{chen2019motion}
Shaoxiang Chen and Yu-Gang Jiang.
\newblock Motion guided spatial attention for video captioning.
\newblock In {\em Proceedings of the AAAI conference on artificial
  intelligence}, volume~33, pages 8191--8198, 2019.

\bibitem{chen2018less}
Yangyu Chen, Shuhui Wang, Weigang Zhang, and Qingming Huang.
\newblock Less is more: Picking informative frames for video captioning.
\newblock In {\em Proceedings of the European conference on computer vision
  (ECCV)}, pages 358--373, 2018.

\bibitem{cornia2020meshed}
Marcella Cornia, Matteo Stefanini, Lorenzo Baraldi, and Rita Cucchiara.
\newblock Meshed-memory transformer for image captioning.
\newblock In {\em Proceedings of the IEEE/CVF Conference on Computer Vision and
  Pattern Recognition}, pages 10578--10587, 2020.

\bibitem{devlin-etal-2019-bert}
Jacob Devlin, Ming-Wei Chang, Kenton Lee, and Kristina Toutanova.
\newblock {BERT}: Pre-training of deep bidirectional transformers for language
  understanding.
\newblock In {\em Proceedings of the 2019 Conference of the North {A}merican
  Chapter of the Association for Computational Linguistics: Human Language
  Technologies, Volume 1 (Long and Short Papers)}, pages 4171--4186,
  Minneapolis, Minnesota, June 2019. Association for Computational Linguistics.

\bibitem{ding2021cogview}
Ming Ding, Zhuoyi Yang, Wenyi Hong, Wendi Zheng, Chang Zhou, Da Yin, Junyang
  Lin, Xu Zou, Zhou Shao, Hongxia Yang, et~al.
\newblock Cogview: Mastering text-to-image generation via transformers.
\newblock {\em arXiv preprint arXiv:2105.13290}, 2021.

\bibitem{donahue2015long}
Jeffrey Donahue, Lisa Anne~Hendricks, Sergio Guadarrama, Marcus Rohrbach,
  Subhashini Venugopalan, Kate Saenko, and Trevor Darrell.
\newblock Long-term recurrent convolutional networks for visual recognition and
  description.
\newblock In {\em Proceedings of the IEEE conference on computer vision and
  pattern recognition}, pages 2625--2634, 2015.

\bibitem{gan2017stylenet}
Chuang Gan, Zhe Gan, Xiaodong He, Jianfeng Gao, and Li Deng.
\newblock Stylenet: Generating attractive visual captions with styles.
\newblock In {\em Proceedings of the IEEE Conference on Computer Vision and
  Pattern Recognition}, pages 3137--3146, 2017.

\bibitem{gan2017semantic}
Zhe Gan, Chuang Gan, Xiaodong He, Yunchen Pu, Kenneth Tran, Jianfeng Gao,
  Lawrence Carin, and Li Deng.
\newblock Semantic compositional networks for visual captioning.
\newblock In {\em Proceedings of the IEEE conference on computer vision and
  pattern recognition}, pages 5630--5639, 2017.

\bibitem{gao2017video}
Lianli Gao, Zhao Guo, Hanwang Zhang, Xing Xu, and Heng~Tao Shen.
\newblock Video captioning with attention-based lstm and semantic consistency.
\newblock {\em IEEE Transactions on Multimedia}, 19(9):2045--2055, 2017.

\bibitem{gu2017empirical}
Jiuxiang Gu, Gang Wang, Jianfei Cai, and Tsuhan Chen.
\newblock An empirical study of language cnn for image captioning.
\newblock In {\em Proceedings of the IEEE International Conference on Computer
  Vision}, pages 1222--1231, 2017.

\bibitem{guo2020normalized}
Longteng Guo, Jing Liu, Xinxin Zhu, Peng Yao, Shichen Lu, and Hanqing Lu.
\newblock Normalized and geometry-aware self-attention network for image
  captioning.
\newblock In {\em Proceedings of the IEEE/CVF Conference on Computer Vision and
  Pattern Recognition}, pages 10327--10336, 2020.

\bibitem{he2016identity}
Kaiming He, Xiangyu Zhang, Shaoqing Ren, and Jian Sun.
\newblock Identity mappings in deep residual networks.
\newblock In {\em European conference on computer vision}, pages 630--645.
  Springer, 2016.

\bibitem{he2020image}
Sen He, Wentong Liao, Hamed~R Tavakoli, Michael Yang, Bodo Rosenhahn, and
  Nicolas Pugeault.
\newblock Image captioning through image transformer.
\newblock In {\em Proceedings of the Asian Conference on Computer Vision},
  2020.

\bibitem{hershey2017cnn}
Shawn Hershey, Sourish Chaudhuri, Daniel~PW Ellis, Jort~F Gemmeke, Aren Jansen,
  R~Channing Moore, Manoj Plakal, Devin Platt, Rif~A Saurous, Bryan Seybold,
  et~al.
\newblock Cnn architectures for large-scale audio classification.
\newblock In {\em 2017 ieee international conference on acoustics, speech and
  signal processing (icassp)}, pages 131--135. IEEE, 2017.

\bibitem{hou2019joint}
Jingyi Hou, Xinxiao Wu, Wentian Zhao, Jiebo Luo, and Yunde Jia.
\newblock Joint syntax representation learning and visual cue translation for
  video captioning.
\newblock In {\em Proceedings of the IEEE/CVF International Conference on
  Computer Vision}, pages 8918--8927, 2019.

\bibitem{johnson2016densecap}
Justin Johnson, Andrej Karpathy, and Li Fei-Fei.
\newblock Densecap: Fully convolutional localization networks for dense
  captioning.
\newblock In {\em Proceedings of the IEEE Conference on Computer Vision and
  Pattern Recognition}, pages 4565--4574, 2016.

\bibitem{karpathy2015deep}
Andrej Karpathy and Li Fei-Fei.
\newblock Deep visual-semantic alignments for generating image descriptions.
\newblock In {\em Proceedings of the IEEE conference on computer vision and
  pattern recognition}, pages 3128--3137, 2015.

\bibitem{kay2017kinetics}
Will Kay, Joao Carreira, Karen Simonyan, Brian Zhang, Chloe Hillier, Sudheendra
  Vijayanarasimhan, Fabio Viola, Tim Green, Trevor Back, Paul Natsev, et~al.
\newblock The kinetics human action video dataset.
\newblock {\em arXiv preprint arXiv:1705.06950}, 2017.

\bibitem{kingma2014adam}
Diederik~P Kingma and Jimmy Ba.
\newblock Adam: A method for stochastic optimization.
\newblock {\em arXiv preprint arXiv:1412.6980}, 2014.

\bibitem{li2019entangled}
Guang Li, Linchao Zhu, Ping Liu, and Yi Yang.
\newblock Entangled transformer for image captioning.
\newblock In {\em Proceedings of the IEEE/CVF International Conference on
  Computer Vision}, pages 8928--8937, 2019.

\bibitem{lin2020multi}
Ke Lin, Zhuoxin Gan, and Liwei Wang.
\newblock Multi-modal feature fusion with feature attention for vatex
  captioning challenge 2020.
\newblock {\em arXiv preprint arXiv:2006.03315}, 2020.

\bibitem{liu2020sibnet}
Sheng Liu, Zhou Ren, and Junsong Yuan.
\newblock Sibnet: Sibling convolutional encoder for video captioning.
\newblock {\em IEEE transactions on pattern analysis and machine intelligence},
  2020.

\bibitem{liu2018show}
Xihui Liu, Hongsheng Li, Jing Shao, Dapeng Chen, and Xiaogang Wang.
\newblock Show, tell and discriminate: Image captioning by self-retrieval with
  partially labeled data.
\newblock In {\em Proceedings of the European Conference on Computer Vision
  (ECCV)}, pages 338--354, 2018.

\bibitem{long2018video}
Xiang Long, Chuang Gan, and Gerard de Melo.
\newblock Video captioning with multi-faceted attention.
\newblock {\em Transactions of the Association for Computational Linguistics},
  6:173--184, 2018.

\bibitem{loshchilov2016sgdr}
Ilya Loshchilov and Frank Hutter.
\newblock Sgdr: Stochastic gradient descent with warm restarts.
\newblock {\em arXiv preprint arXiv:1608.03983}, 2016.

\bibitem{pan2020auto}
Yingwei Pan, Yehao Li, Jianjie Luo, Jun Xu, Ting Yao, and Tao Mei.
\newblock Auto-captions on gif: A large-scale video-sentence dataset for
  vision-language pre-training.
\newblock {\em arXiv preprint arXiv:2007.02375}, 2020.

\bibitem{pan2016jointly}
Yingwei Pan, Tao Mei, Ting Yao, Houqiang Li, and Yong Rui.
\newblock Jointly modeling embedding and translation to bridge video and
  language.
\newblock In {\em Proceedings of the IEEE conference on computer vision and
  pattern recognition}, pages 4594--4602, 2016.

\bibitem{pan2017video}
Yingwei Pan, Ting Yao, Houqiang Li, and Tao Mei.
\newblock Video captioning with transferred semantic attributes.
\newblock In {\em Proceedings of the IEEE conference on computer vision and
  pattern recognition}, pages 6504--6512, 2017.

\bibitem{pan2020x}
Yingwei Pan, Ting Yao, Yehao Li, and Tao Mei.
\newblock X-linear attention networks for image captioning.
\newblock In {\em Proceedings of the IEEE/CVF Conference on Computer Vision and
  Pattern Recognition}, pages 10971--10980, 2020.

\bibitem{papineni2002bleu}
Kishore Papineni, Salim Roukos, Todd Ward, and Wei-Jing Zhu.
\newblock Bleu: a method for automatic evaluation of machine translation.
\newblock In {\em Proceedings of the 40th annual meeting of the Association for
  Computational Linguistics}, pages 311--318, 2002.

\bibitem{pei2019memory}
Wenjie Pei, Jiyuan Zhang, Xiangrong Wang, Lei Ke, Xiaoyong Shen, and Yu-Wing
  Tai.
\newblock Memory-attended recurrent network for video captioning.
\newblock In {\em Proceedings of the IEEE/CVF Conference on Computer Vision and
  Pattern Recognition}, pages 8347--8356, 2019.

\bibitem{perez2021improving}
Jesus Perez-Martin, Benjamin Bustos, and Jorge P{\'e}rez.
\newblock Improving video captioning with temporal composition of a
  visual-syntactic embedding.
\newblock In {\em Proceedings of the IEEE/CVF Winter Conference on Applications
  of Computer Vision}, pages 3039--3049, 2021.

\bibitem{press2017using}
Ofir Press and Lior Wolf.
\newblock Using the output embedding to improve language models.
\newblock In {\em Proceedings of the 15th Conference of the European Chapter of
  the Association for Computational Linguistics: Volume 2, Short Papers}, pages
  157--163, 2017.

\bibitem{ranzato2015sequence}
Marc'Aurelio Ranzato, Sumit Chopra, Michael Auli, and Wojciech Zaremba.
\newblock Sequence level training with recurrent neural networks.
\newblock {\em arXiv preprint arXiv:1511.06732}, 2015.

\bibitem{rennie2017self}
Steven~J Rennie, Etienne Marcheret, Youssef Mroueh, Jerret Ross, and Vaibhava
  Goel.
\newblock Self-critical sequence training for image captioning.
\newblock In {\em Proceedings of the IEEE Conference on Computer Vision and
  Pattern Recognition}, pages 7008--7024, 2017.

\bibitem{singh2020nits}
Alok Singh, Thoudam~Doren Singh, and Sivaji Bandyopadhyay.
\newblock Nits-vc system for vatex video captioning challenge 2020.
\newblock {\em arXiv preprint arXiv:2006.04058}, 2020.

\bibitem{sun2019videobert}
Chen Sun, Austin Myers, Carl Vondrick, Kevin Murphy, and Cordelia Schmid.
\newblock Videobert: A joint model for video and language representation
  learning.
\newblock In {\em Proceedings of the IEEE/CVF International Conference on
  Computer Vision}, pages 7464--7473, 2019.

\bibitem{vaswani2017attention}
Ashish Vaswani, Noam Shazeer, Niki Parmar, Jakob Uszkoreit, Llion Jones,
  Aidan~N Gomez, {\L}ukasz Kaiser, and Illia Polosukhin.
\newblock Attention is all you need.
\newblock In {\em Advances in neural information processing systems}, pages
  5998--6008, 2017.

\bibitem{vedantam2015cider}
Ramakrishna Vedantam, C Lawrence~Zitnick, and Devi Parikh.
\newblock Cider: Consensus-based image description evaluation.
\newblock In {\em Proceedings of the IEEE conference on computer vision and
  pattern recognition}, pages 4566--4575, 2015.

\bibitem{vinyals2015show}
Oriol Vinyals, Alexander Toshev, Samy Bengio, and Dumitru Erhan.
\newblock Show and tell: A neural image caption generator.
\newblock In {\em Proceedings of the IEEE conference on computer vision and
  pattern recognition}, pages 3156--3164, 2015.

\bibitem{wang2019controllable}
Bairui Wang, Lin Ma, Wei Zhang, Wenhao Jiang, Jingwen Wang, and Wei Liu.
\newblock Controllable video captioning with pos sequence guidance based on
  gated fusion network.
\newblock In {\em Proceedings of the IEEE/CVF International Conference on
  Computer Vision}, pages 2641--2650, 2019.

\bibitem{wang2018reconstruction}
Bairui Wang, Lin Ma, Wei Zhang, and Wei Liu.
\newblock Reconstruction network for video captioning.
\newblock In {\em Proceedings of the IEEE conference on computer vision and
  pattern recognition}, pages 7622--7631, 2018.

\bibitem{wang2018m3}
Junbo Wang, Wei Wang, Yan Huang, Liang Wang, and Tieniu Tan.
\newblock M3: Multimodal memory modelling for video captioning.
\newblock In {\em Proceedings of the IEEE conference on computer vision and
  pattern recognition}, pages 7512--7520, 2018.

\bibitem{wang2019vatex}
Xin Wang, Jiawei Wu, Junkun Chen, Lei Li, Yuan-Fang Wang, and William~Yang
  Wang.
\newblock Vatex: A large-scale, high-quality multilingual dataset for
  video-and-language research.
\newblock In {\em Proceedings of the IEEE/CVF International Conference on
  Computer Vision}, pages 4581--4591, 2019.

\bibitem{wu2016google}
Yonghui Wu, Mike Schuster, Zhifeng Chen, Quoc~V Le, Mohammad Norouzi, Wolfgang
  Macherey, Maxim Krikun, Yuan Cao, Qin Gao, Klaus Macherey, et~al.
\newblock Google's neural machine translation system: Bridging the gap between
  human and machine translation.
\newblock {\em arXiv preprint arXiv:1609.08144}, 2016.

\bibitem{xu2016msr}
Jun Xu, Tao Mei, Ting Yao, and Yong Rui.
\newblock Msr-vtt: A large video description dataset for bridging video and
  language.
\newblock In {\em Proceedings of the IEEE conference on computer vision and
  pattern recognition}, pages 5288--5296, 2016.

\bibitem{xu2015show}
Kelvin Xu, Jimmy Ba, Ryan Kiros, Kyunghyun Cho, Aaron Courville, Ruslan
  Salakhudinov, Rich Zemel, and Yoshua Bengio.
\newblock Show, attend and tell: Neural image caption generation with visual
  attention.
\newblock In {\em International conference on machine learning}, pages
  2048--2057. PMLR, 2015.

\bibitem{yu2019multimodal}
Jun Yu, Jing Li, Zhou Yu, and Qingming Huang.
\newblock Multimodal transformer with multi-view visual representation for
  image captioning.
\newblock {\em IEEE transactions on circuits and systems for video technology},
  30(12):4467--4480, 2019.

\bibitem{zhang2019object}
Junchao Zhang and Yuxin Peng.
\newblock Object-aware aggregation with bidirectional temporal graph for video
  captioning.
\newblock In {\em Proceedings of the IEEE/CVF Conference on Computer Vision and
  Pattern Recognition}, pages 8327--8336, 2019.

\bibitem{zhang2021open}
Ziqi Zhang, Zhongang Qi, Chunfeng Yuan, Ying Shan, Bing Li, Ying Deng, and
  Weiming Hu.
\newblock Open-book video captioning with retrieve-copy-generate network.
\newblock In {\em Proceedings of the IEEE/CVF Conference on Computer Vision and
  Pattern Recognition}, pages 9837--9846, 2021.

\bibitem{zhang2020object}
Ziqi Zhang, Yaya Shi, Chunfeng Yuan, Bing Li, Peijin Wang, Weiming Hu, and
  Zheng-Jun Zha.
\newblock Object relational graph with teacher-recommended learning for video
  captioning.
\newblock In {\em Proceedings of the IEEE/CVF conference on computer vision and
  pattern recognition}, pages 13278--13288, 2020.

\bibitem{zhu2019vatex}
Xinxin Zhu, Longteng Guo, Peng Yao, Shichen Lu, Wei Liu, and Jing Liu.
\newblock Vatex video captioning challenge 2020: Multi-view features and hybrid
  reward strategies for video captioning.
\newblock {\em arXiv preprint arXiv:1910.11102}, 2019.

\end{thebibliography}
